\pdfoutput=1

\documentclass[11pt]{article}

\usepackage[final]{acl}

\usepackage{times}
\usepackage{latexsym}

\usepackage{booktabs}

\usepackage{paralist}

\usepackage[most]{tcolorbox}

\usepackage[T1]{fontenc}

\usepackage[utf8]{inputenc}

\usepackage{microtype}

\usepackage{inconsolata}

\usepackage{graphicx}

\newcommand{\remove}[1]{}

\title{Reliable Proof Generation with LLMs via Analogical Retrieval and Symbolic Verification: A Case Study in Euclidean Geometry}

\author{
\textbf{Oren Sultan}, \textbf{Eitan Stern}, \textbf{Dafna Shahaf} \\
The Hebrew University of Jerusalem \\
\texttt{\{oren.sultan,eitan.stern,dshahaf\}@cs.huji.ac.il} \\
}

\begin{document}
\maketitle

\begin{abstract}

Large language models (LLMs) struggle with formal domains that require rigorous logical deduction and symbolic reasoning, such as mathematical proof generation. 
We propose a neuro-symbolic approach centered on the hypothesis that structurally analogous problems often admit similar proofs. 
As a proof-of-concept, we focus on SAT-level geometry problems.
%
%
%
%
Our approach is two-fold: 
(1) We retrieve \emph{analogous problems} and use their proofs to guide the LLM, and 
(2) a \emph{formal verifier} evaluates the generated proofs and provides feedback, helping the model fix incorrect proofs.

Our complete pipeline substantially improves proof accuracy across model families, achieving 68\%-96\% accuracy compared with 10\%-44\% for LLM-only baselines that use neither analogy retrieval nor verifier feedback. When comparing against models with the same verifier feedback and inference budget, analogical guidance improves accuracy  from 88\% to 96\% for GPT-5, 78\% to 86\% for Claude Sonnet 4.6, 72\% to 86\% for Gemini-Flash-2.5, and 52\% to 80\% for OpenAI o1.
More broadly, shifting to LLMs that generate provably correct conclusions has the potential to dramatically improve their reliability, accuracy and consistency, unlocking complex tasks and critical real-world applications that require trustworthiness.






\end{abstract}

\section{Introduction}

\begin{figure*}[t]
\includegraphics[width=0.9\textwidth]{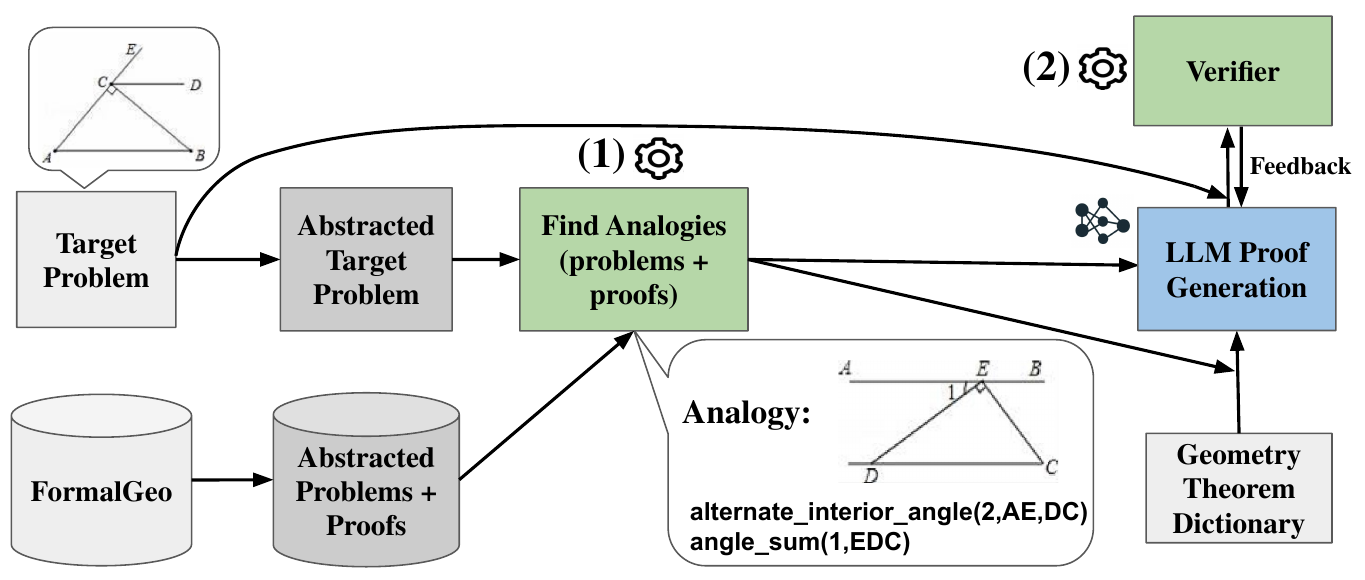}
\centering
\caption{\textbf{Our neuro-symbolic approach.} Given a target problem from the FormalGeo-7k dataset, we first convert it into an \emph{abstract form} by replacing entity names (e.g., lines, angles) and specific numeric values with placeholders (~\S\ref{subsec:problems_abstraction}).  
We then \emph{retrieve structurally similar problems} from the abstracted dataset by computing Jaccard similarity over key formal components: \textbf{construction} (entities and geometric relations), \textbf{conditions} (e.g., angle equalities, segment lengths), and \textbf{goal} (the conclusion to be proven). This is based on the observation that structurally similar problems often share proof patterns (~\S\ref{subsec:similar_problems_retrieval}).
The retrieved problems, along with their corresponding formal proofs, are presented to an LLM as \emph{in-context examples}, together with the available theorems from the Geometry Theorem Dictionary, to guide \emph{proof generation} for the target problem (~\S\ref{subsec:llm_solver}).  
Finally, a \emph{symbolic verifier} iteratively checks the generated proof and provides feedback until a correct proof is produced or a retry limit is reached (~\S\ref{subsec:verifier}).
}
\label{fig:pipeline}
\end{figure*}

\begin{figure*}[t]
\centering
\includegraphics[width=0.9\textwidth]{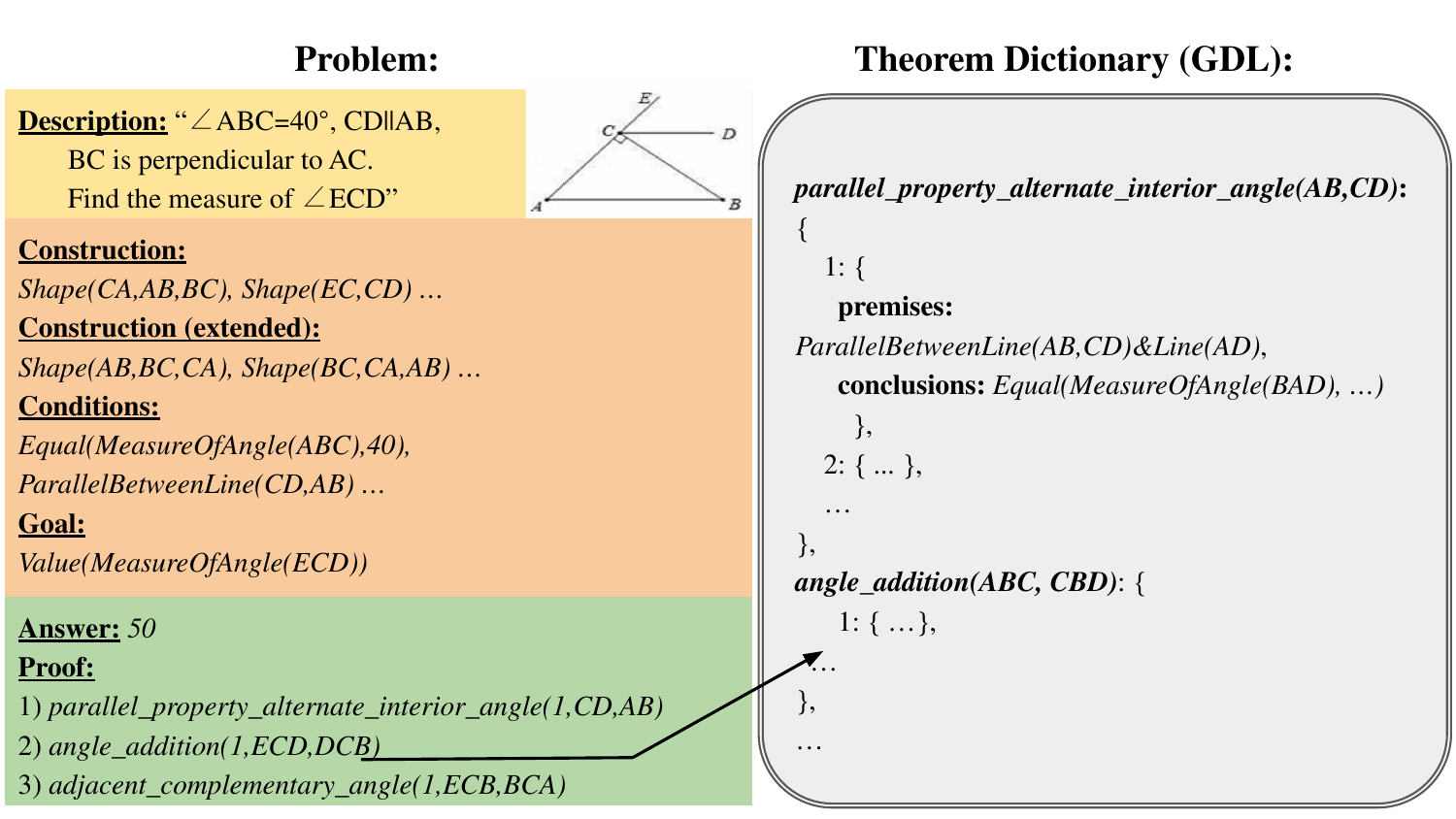}
\caption{An example problem from FormalGeo-7k.  
\textbf{Left}: Problem inputs, including a natural-language description and formal representations: construction (entities and relations), extended construction (inferred via extension rules), conditions, and goal. The input also includes a numeric answer (or expression) and a formal proof, with each step invoking a theorem from the dictionary.
\textbf{Right}: The Theorem Dictionary, with theorem IDs, arguments, premises, and conclusions. The dataset also includes diagrams (which we do not use), shown here for illustration.
}
\label{fig:dataset_example}
\end{figure*}


LLMs have considerable potential for mathematical proof generation \cite{tafjord2020proofwriter, yasunaga2023large, trinh2024solving, ren2025deepseek}.
Unlike symbolic theorem provers, LLMs can directly process natural-language problem descriptions, draw on broad mathematical knowledge acquired during pretraining, and flexibly explore possible proof strategies. 
However, despite substantial progress in LLM capabilities, their generated proofs remain unreliable without verification \cite{cobbe2021training, lightman2024let}.

Mathematical 
proofs demand rigorous logical deduction, symbolic manipulation, and an understanding 
of abstract concepts that go beyond statistical correlations in language. The requirement for absolute 
truth and the absence of ambiguity in mathematical reasoning present a significant challenge for models trained to generate \emph{plausible} text rather than formally valid inferences \cite{singh2024exposing, pan2025lemma}. 
In addition, the often lengthy nature of proofs necessitates a level of sustained logical coherence and hierarchical reasoning 
that is hard for current LLMs. Recent work has shown that altering even superficial aspects of mathematical problems results in significant performance drops \cite{mirzadeh2024gsm}, suggesting that their success often hinges on pattern matching rather than genuine mathematical reasoning.

The lengthy nature of proofs also implies that the space of possible proofs can get prohibitively large, and general-purpose LLMs often struggle to search through it. We address this challenge through analogy; our central hypothesis is that \emph{structurally analogous geometry problems admit similar proofs}.

In this work, we introduce a neuro-symbolic approach that combines the generative strengths of LLMs with two complementary structured components: (1) \textbf{analogical guidance} and (2) \textbf{symbolic verification}. See Figure~\ref{fig:pipeline} for an illustration.

The first component retrieves analogous problems and their proofs to guide the model. This is inspired by cognitive science, where analogy is recognized as fundamental to human problem-solving and generalization \cite{gentner1983structure, holyoak1996mental}, as well as recent work showing that prompting LLMs to generate analogous problems and solutions can improve performance on math problems \cite{yasunaga2023large}.

 Since the target proof is unavailable at inference time, we construct abstract symbolic representations of problem statements and learn to predict proof similarity from these features. Retrieved proofs of analogous problems then provide useful starting points for proof generation and help identify the most relevant theorems to present to the model, substantially reducing costs.

To complement this, our second component employs a symbolic verifier that checks the generated proofs and provides structured feedback. This feedback drives an iterative loop, enabling the model to revise its output until a valid proof is obtained. 

As a proof of concept, we focus on Euclidean geometry -- a symbolic, verifiable domain rich in structural analogies.
Our goal is to evaluate whether our symbolic augmentations improve the proof-generation capabilities of \emph{general-purpose LLMs}. Thus, while we use geometry as our testbed, our focus is \emph{not} on competing with specialized state-of-the-art geometry solvers, but rather on quantifying the gains enabled by our method.



Enabling LLMs to generate formally verifiable proofs can improve their reliability,\footnote{Throughout this paper, we use \emph{reliable} to refer to the generation of proofs formally verified by a symbolic verifier.} in applications across math, science and education \cite{welleck2022naturalprover, gupta2025beyond, kumar2023math}.

\textbf{Our main contributions are:}
\textbf{(1)} We propose a neuro-symbolic system that aids LLMs in proof generation by providing analogical guidance and verification feedback.
\textbf{(2)} We design a symbolic verifier tailored to geometry proofs with expressive feedback. In contrast to other works, we evaluate the {\bf entire proof}, not just the final numeric answer.
\textbf{(3)} Our method substantially improves proof accuracy across all models while reducing computational cost through focused context construction. Our complete pipeline achieves 68\%-96\% accuracy, compared with 10\%-44\% for LLM-only baselines. Under matched inference budgets with verifier feedback, analogical guidance improves accuracy from 88\% to 96\% for GPT-5, 78\% to 86\% for Claude Sonnet 4.6, 72\% to 86\% for Gemini-Flash-2.5, and 52\% to 80\% for o1. At the same time, it substantially reduces the theorem dictionary size (e.g., from 18K to 2.5K tokens on average for o1).
\textbf{(4)} We release data and code at: \url{https://github.com/orensul/LLMs_proof_generation}.
 



\section{Problem Formulation}
\label{sec:problem_defintion}



We demonstrate our ideas in the domain of Euclidean geometry. Our input is a geometry problem, described in both natural language and via a formal representation. The description includes the geometric entities involved (e.g., lines, angles), their relationships (e.g., perpendicular, collinear),
and measurements or algebraic expressions over them. We also receive a \emph{goal}, some quantity to be determined (e.g., the length of a line). In addition, the model has access to a dictionary of theorems that may be used in the proof.

The output is a formal proof that derives the goal from the given conditions and theorems, along with the final, numeric answer. The proof consists of steps, each applying a specific theorem from the dictionary. See Figure~\ref{fig:dataset_example} for an example.

\section{Approach}

Our goal is to develop a system that assists LLMs in proof generation.
As a proof of concept, we focus on geometry problems from the FormalGeo-7k dataset~\cite{zhang2023formalgeo}, containing 6,981 SAT-level Euclidean geometry problems.


Our approach is two-fold (see Figure~\ref{fig:pipeline}). Given a target problem, we first 
build on the insight that structurally similar problems often admit similar proofs. We abstract all problems in the dataset {(\S\ref{subsec:problems_abstraction})}, retrieve {\bf analogous problems} {(similar to the target on an abstract level, \S\ref{subsec:similar_problems_retrieval})}, and use the top-ranked analogies and their corresponding proofs as few-shot examples 
{(\S\ref{subsec:llm_solver})}.
Next, we employ a {\bf symbolic verifier} that iteratively provides feedback on the validity of the generated proofs {(\S\ref{subsec:verifier})}.

\subsection{Problem Abstraction}
\label{subsec:problems_abstraction}
Analogous problems share a similar underlying structure, but could differ on surface-level details such as entity names or measurements.
To identify such problems, the pipeline first abstracts the target problem and all problems in the dataset. 
We chose a very simple abstraction schema:
 entity names are replaced with ``<word>'' and numbers with ``<num>''.  
 For example,  
 ``Equal(MeasureOfAngle(ABC),40)'' becomes ``Equal(MeasureOfAngle(<word>), <num>)''. This process is applied to the formal representations of construction, conditions, and goal. More nuanced abstraction schemas (e.g., preserving shared symbols between words) are left for future work.



\subsection{Analogous Problems Retrieval}
\label{subsec:similar_problems_retrieval}



Our goal in this section is to retrieve problems whose (known) proof is similar to the (unknown) target proof. We conjecture that providing the model with these proof examples in context will improve its ability to generate a correct proof.

Our underlying working hypothesis is that \emph{similar problems often yield similar proofs}.
%
%
More concretely, we posit we can identify problems with potentially useful proofs by finding analogous problems (i.e., problems which are structurally similar to the target problem). To do this, we train a regressor to predict proof similarity between two problems, based on their structural similarity.



\noindent\textbf{Dataset and Training.}  
For each pair of abstracted problems, we compute the Jaccard similarity over the multi-sets of \emph{construction} and \emph{condition}  representations. We define \emph{goal} similarity as 1 if the goals match exactly and 0 otherwise. These three features form the input to our model. Abstract proof similarity, computed again via Jaccard similarity, serves as the label.

The dataset contains $\sim24.4$M  problem pairs (all combinations of 6,981 problems). Proof similarity scores are binned into five intervals of width 0.2. 
As the distribution is highly imbalanced, with 88\% of pairs in the first bin (0-0.2), we down-sample all bins to the size of the smallest (131K, 0.8-1), yielding a balanced dataset of 655K pairs. We split the data into 90\% training and 10\% evaluation.

We train a simple three-layer neural network 
using mean squared error (MSE) loss and the Adam optimizer, with a learning rate of 0.001, batch size of 32, and 10 epochs. See Appendix~\ref{appendix:analogous_problems_retrieval} for details.

\noindent\textbf{Evaluation.}
Note that for our use case, we are only interested in whether we can predict very high proof similarity. 
We select all pairs with predicted similarity above 0.95 and measure the fraction whose ground-truth proof similarity is in the top two bins. While only 1.28\% of  pairs exceed this threshold, 
71\% of our predictions do.

We are encouraged by the results, which confirm that our regressor can identify proofs similar to the (unknown) target proof, based on the target problem's description alone.


\subsection{LLM Proof Generation}
\label{subsec:llm_solver}

\begin{table}[!t]
    \centering
    \small
    \setlength{\tabcolsep}{4pt} 
    \begin{tabular}{l cc cc}
        \toprule
        \textbf{$k$} & \multicolumn{2}{c}{\textbf{Analogy}} & \multicolumn{2}{c}{\textbf{Random}} \\
        & Coverage~$\uparrow$ & Theorems~$\downarrow$ & Coverage~$\uparrow$  & Theorems~$\downarrow$  \\
        \midrule
        20 & 88\% & 11.06 & 62\% & 32.92 \\
        \midrule
        50 & 93\% & 18.57 & 79\% & 49.87 \\
        \midrule
        100 & 96\% & 26.66 & 88\% & 66.98 \\
        \midrule
        150 & 98\% & 31.96 & 92\% & 82.82 \\
        \bottomrule
    \end{tabular}
    \caption{Average number of problems whose entire proof was covered by the union of theorems from their top‑$k$ analogies vs.~random problems, over 100 random target problems (20 per level, 1–5). 
    Analogies consistently achieve higher coverage despite fewer theorems.}
    \label{tab:theorems_coverage}
\end{table}




We use a few-shot prompt (in-context learning) that begins with the target problem, including its textual description, construction, conditions and goal. This is followed by the top analogous problems selected using the regressor from Section~\ref{subsec:similar_problems_retrieval}, along with their full proofs and numeric answers.
Additionally, the LLM is given a theorem dictionary, which defines geometry theorems in terms of formal premises and conclusions (see Figure~\ref{fig:dataset_example}). The LLM's task is to generate a correct proof, using only the theorems from the dictionary. See Appendix~\ref{appendix:LLM_proof_generation} for the prompt.

One challenge we encountered is that the full theorem dictionary contains 196 theorems (234 including  variations), resulting in a large token count and, consequently, 
high costs. This also limits scalability, as adding more theorems would further increase the input size.

To address this problem, we propose a more efficient approach. We have just shown that analogous problems tend to have similar proofs; thus, we test whether we could similarly narrow down the dictionary to include only \emph{theorems used in analogous proofs}. 
This reduction not only reduces costs, but also narrows the search space, helping the model focus on more relevant theorems. 

To evaluate the effectiveness of this approach, we measure the extent to which the narrowed-down dictionary still captures all the theorems needed for the proof of the target problem.
We sample 100 problems (20 from each level 1-5) 
and evaluate how many of their target proofs are completely covered using theorems from their top-$k$ similar problems (retrieved with the regressor of  Section~\ref{subsec:similar_problems_retrieval}), and compare to a random set of $k$ problems.

Note that even if a proof of a problem includes a theorem not present in its analogous set, it might still be possible to construct a valid proof using only the covered theorems. 
Therefore, our coverage metric is a conservative estimate of effectiveness.

See Table~\ref{tab:theorems_coverage} for results.  
Theorems from analogous problems consistently outperform the baseline of theorems from random $k$ problems, providing higher coverage despite lower number of theorems. A larger $k$ increases coverage but also expands the input; we find that $k=100$ offers a good trade-off, achieving coverage of 96\% with only 26.66 theorems on average, that is 13.6\% of the full dictionary. Exploring other values of $k$ is left for future work.

\subsection{Verifier Iterative Feedback}
\label{subsec:verifier}


We next endow the LLM with an external verifier that can guide it through a feedback loop. 
This is inspired by similar efforts showing that code-writing LLMs can improve with iterative corrections 
\cite{peng2025perfcodegen, palavalli2024using}.
After each attempt to produce a proof, the verifier provides natural-language feedback, specifying the first error found in the proof. The LLM incorporates it into its context for the next iteration. The loop continues until the proof is verified or the maximum number of retries is reached (see Section~\ref{subsec:setup}).



The verifier is a symbolic reasoning system capable of performing both formal logic checks and algebraic reasoning. 
Internally, it encodes proof steps and geometric constraints as logical formulas and algebraic expressions, and evaluates them using satisfiability modulo theories (SMT). 

Importantly, the verifier does not know the solution. Instead, it assesses whether the numerical answer is  entailed by the constraints imposed by the proof. If the proof is valid and the answer can indeed be inferred from it, the proof is accepted. 

\noindent\textbf{Error tiers.}
To analyze where the LLM struggles, we define three verifier-identified error tiers:

\begin{compactenum}
    \item \textbf{Theorem call syntax violation}: Syntax errors. Common issues include undefined theorems  or  incorrect argument signatures. 

    \item \textbf{Premise violation}: Theorem calls that rely on premises that have not been derived from the problem description or preceding proof steps. 
    Feedback identifies the missing premise and lists all premises derived so far.

    \item \textbf{Goal not reached}: The proof contains no errors but does not reach the goal.  
    Feedback indicates that either (1) the proof is underconstrained and multiple solutions exist, or (2) that the (unique) solution derived by the verifier differs from the LLM's answer.
\end{compactenum}
\noindent See Appendix~\ref{appendix:verifier_details} for examples of error messages from the different tiers. 

\noindent\textbf{Implementation details.}
To identify tier-1 errors, we extended the  implementation of \citet{zhang2023formalgeo}. 
To identify tier-2 and tier-3 errors we use the Z3 Theorem Prover~\cite{de2008z3}, a state-of-the-art SMT solver, that encodes algebraic constraints derived from geometric properties and verifies their logical consistency.

We also augment Z3 with symbolic workarounds for trigonometric functions it does not natively support.
See Appendix~\ref{appendix:verifier_details} for more details.




\section{Experimental Setup}

\label{subsec:research_questions}
\label{subsec:setup}

We evaluate the performance of our method on the FormalGeo-7K dataset. Our main research questions are as follows:

\noindent\textbf{RQ1:} Does our method lead to a higher rate of correct proofs generated by the LLM?

\noindent\textbf{RQ2 (Ablation):} What is the individual contribution of different components in our method?


\noindent\textbf{A Note on Baselines for Comparison.}
Our goal is to measure how symbolic augmentation improves formal proof generation in \emph{existing LLMs}, rather than compete with specialized geometry solvers. Similarly, we evaluate whether analogy retrieval improves proof generation, rather than optimize the retrieval method itself. We evaluate retrieval in three ways: \emph{intrinsic retriever evaluation} via proof-similarity prediction (\S\ref{subsec:similar_problems_retrieval}); \emph{intrinsic comparison to random retrieval} via theorem coverage (Table~\ref{tab:theorems_coverage}); and \emph{extrinsic comparison to random retrieval} via downstream proof accuracy under matched inference budgets (Table~\ref{tab:main_results_tab}). Comparisons with additional retrieval methods are left to future work.


We now share details on our experimental setup.

\noindent\textbf{Input problems.}
We randomly sample 10 problems for each difficulty level from 1 to 5 from the FormalGeo-7K dataset, resulting in a total of 50 problems. Levels correspond to the number of steps in the ground-truth proof. 

\noindent\textbf{Base models.} 
We evaluate \emph{reasoning-capable} models from multiple families using recommended configurations. 
We evaluate OpenAI's o1~\cite{jaech2024openai}, GPT-5~\cite{gpt-5}, Gemini-2.5-Flash~\cite{comanici2025gemini} and Claude Sonnet 4.6~\cite{claude_sonnet_4.6}. Throughout the paper, \emph{baseline} refers to the non-augmented variant of each of these models (i.e., without our neuro-symbolic method), rather than to any specific model.

We use OpenAI o1 with the default settings; GPT-5 with medium \emph{reasoning}; Gemini-2.5-Flash with \emph{t=0.7}; and Claude Sonnet 4.6 with \emph{t=1.0}. Full configuration details are provided in our codebase.

We additionally evaluated \emph{non-reasoning} models, including GPT-4 and GPT-4o~\cite{achiam2023gpt, hurst2024gpt}, and found that they frequently generate proofs with numerous syntax errors. 
Smaller open-weight models such as Qwen-3-8B and Qwen-3-14B~\cite{yang2025qwen3} similarly struggle to adhere to the required output format.

\noindent\textbf{Few-shot variants.}  
For each base model, we evaluate two variants: an analogy-based variant using our method, and a baseline variant without it.
\begin{compactitem}
  \item \textbf{Analogy-based:} We provide $k$ similar (analogous) problems and their proofs as few-shot examples, along with an abridged theorem dictionary which includes only theorems from the proofs of similar problems (Section~\ref{subsec:llm_solver}).
  \item \textbf{Base model (non-analogy):} We provide $k$ random problems and their proofs as few-shot examples, along with the complete dictionary.
\end{compactitem}



\noindent {\bf A note about $k$.} In our preliminary exploration, no model was able to correctly follow the proof format with zero-shot prompts (unsurprisingly). We have found that $k=5$ works well in practice; testing the effect of $k$ in more depth is left for future work.

\noindent {\bf Feedback iterations and multiple runs.}
For each base model and variant (analogy-based and non-analogy), we evaluate several inference settings involving verifier-guided feedback and repeated runs.
As outlined in Section~\ref{subsec:verifier}, we integrate the LLM into a \emph{feedback loop} guided by a verifier, allowing the LLM up to five iterations (retries following feedback) per run.  
In addition, since LLMs sometimes fail to progress toward the correct solution (even with feedback), we allow restarting from scratch up to three times per problem.
Within each run, the conversation is append-only: failed attempts and verifier feedback are retained, and each retry conditions on the full interaction history within that run. The $n$ runs are independent, each starting from a fresh context.


Overall, we evaluate four base models, each under two prompting variants and three inference settings: single-pass, verifier-guided feedback iterations, and multiple independent runs with feedback iterations. 
In total, this yields 24 configurations.

We denote the different configurations by $\text{base}_{i,j}$ or $\text{ours}_{i,j}$,
where $i$ is the number of runs and $j$ the number of verifier-guided retries.

\noindent {\bf Metric.}
We report the fraction of problems for which the model produces a correct proof, for the following settings:
\begin{compactitem}
    \item \textbf{First run, no retries (\%):} The model produces a correct proof on its first attempt.
    \item \textbf{First run, with retries (\%):} Any of $m$ retries (following feedback) in the first run succeeds.
    \item \textbf{Multiple runs, with retries (\%):}  Any attempt across $n$ independent runs (each with up to $m$ retries) succeeds.
\end{compactitem}

More runs and retries increase costs; we found $m=5$, $n=3$ to provide a good tradeoff between performance and budget, and analyze their effect in the next section.
Based on logged API calls, token usage, and current pricing, evaluating 50 problems costs approximately \$100 (o1), \$3 (GPT-5), \$1 (Gemini-2.5-Flash), and \$10 (Claude Sonnet 4.6). The baseline is approximately 3$\times$ more expensive, since it prompts with the full $\sim$18K-token theorem dictionary rather than our $\sim$2.5K-token analogy-derived dictionary, resulting in substantially larger prompts and typically more verifier-guided retries.


\section{Results}
\label{subsec:results}

\begin{table*}[t]
\centering
\setlength{\tabcolsep}{5pt}
\begin{tabular}{l | ccc | ccc | ccc}
\toprule
& \multicolumn{3}{c}{\textbf{First run, no retries}} 
& \multicolumn{3}{c}{\textbf{First run, w/ retries}} 
& \multicolumn{3}{c}{\textbf{Multiple runs, w/ retries}} \\
\cmidrule(lr){2-4} \cmidrule(lr){5-7} \cmidrule(lr){8-10}
\textbf{Model }
& \textit{base}$_{1,1}$ & \textit{ours}$_{1,1}$ & $\Delta$ 
& \textit{base}$_{1,5}$ & \textit{ours}$_{1,5}$ & $\Delta$ 
& \textit{base}$_{3,5}$ & \textit{ours}$_{3,5}$ & $\Delta$ \\
\midrule
o1 
& 10\% & \textbf{48\%} & +38\% 
& 38\% & \textbf{68\%} & +30\% 
& 52\% & \textbf{80\%} & +28\% \\

GPT-5 
& 44\% & \textbf{64\%} & +20\% 
& 80\% & \textbf{96\%} & +16\% 
& 88\% & \textbf{96\%} & +8\% \\

Gemini-Flash-2.5 
& 22\% & \textbf{48\%} & +26\% 
& 58\% & \textbf{74\%} & +16\% 
& 72\% & \textbf{86\%} & +14\% \\

Claude Sonnet-4.6 
& 28\% & \textbf{60\%} & +32\% 
& 58\% & \textbf{78\%} & +20\% 
& 78\% & \textbf{86\%} & +8\% \\
\bottomrule
\end{tabular}
\caption{Proof accuracy (\%) of the base model and our analogy-based method across settings. \textit{base}$_{i,j}$ denotes the base model with up to $i$ runs and $j$ retries, while \textit{ours}$_{i,j}$ denotes our method. $\Delta$ indicates absolute improvement (percentage points). \textbf{Our method consistently improves proof accuracy across all models and settings}.}
\label{tab:main_results_tab}
\end{table*}

Table~\ref{tab:main_results_tab} shows the results for the different settings. 

\subsection{RQ1: Performance gains of our approach.}
We first compare our full pipeline, \textit{ours}$_{3,5}$, against the minimal \textit{base}$_{1,1}$ baseline. Our method substantially improves accuracy across all models: from 10\% to 80\% for o1, 22\% to 86\% for Gemini, 28\% to 86\% for Claude, and 44\% to 96\% for GPT-5.

To isolate the contribution of analogical guidance, we compare \textit{base}$_{3,5}$ and \textit{ours}$_{3,5}$, which use the same inference budget and verifier feedback and differ only in analogical guidance. Accuracy still improves across all models: from 52\% to 80\% for o1, 72\% to 86\% for Gemini, 78\% to 86\% for Claude, and 88\% to 96\% for GPT-5.

Finally, we evaluate our method with a single run (\textit{ours}$_{1,5}$), isolating the benefit from multiple runs. Even without multiple runs, accuracy reaches 68\% for o1, 74\% for Gemini, 78\% for Claude, and 96\% for GPT-5, compared with 10\%, 22\%, 28\%, and 44\% for \textit{base}$_{1,1}$, respectively. Overall, these results show substantial gains from our approach across models and inference settings. See Appendix~\ref{appendix:results_rq1} for detailed results and significance tests.

\subsection{RQ2: Ablations.}
We assess each of our method’s components: analogy retrieval, the verifier, and multiple runs.

\noindent\textbf{Analogy retrieval.}
To isolate the effect of analogy retrieval, we compare our method to the base model under three settings: 
(1) single run no retries, 
(2) single run with verifier-based retries, and 
(3) multiple runs with verifier-based retries.
In setting (1), our method substantially improves over the baseline: o1 increases from 10\% to 48\% accuracy, Gemini from 22\% to 48\%, Claude from 28\% to 60\%, and GPT-5 from 44\% to 64\% ( \textit{base}$_{1,1}$ and \textit{ours}$_{1,1}$).
In setting (2), our method again improves the baseline: o1 increases from 38\% to 68\%, Gemini from 58\% to 74\%, Claude from 58\% to 78\%, and GPT-5 from 80\% to 96\% (\textit{base}${1,5}$ and \textit{ours}${1,5}$).
In setting (3), our method continues to outperform the baseline: o1 improves from 52\% to 80\% accuracy, Gemini from 72\% to 86\%, and Claude from 78\% to 86\%, while GPT-5 from 88\% to 96\% (\textit{base}${3,5}$ and \textit{ours}${3,5}$).
Overall, analogy retrieval consistently improves performance across all settings. See Appendix~\ref{appendix:results_rq2} for a detailed analysis, and statistical significance tests.

The comparison above varies both example selection (retrieved vs.\ random) and theorem dictionary (analogy-derived vs.\ full). To isolate their contributions, we evaluate three intermediate configurations (Table~\ref{tab:ablation_dict}): (i) retrieved + full dictionary, (ii) random + narrowed dictionary, and (iii) retrieved + random size-matched dictionary. 

Our full method performs best across all inference budgets; (i) and (iii) outperform (ii), demonstrating the benefit of analogy retrieval, while our gains over (i) and (iii) demonstrate the additional benefit of analogy-derived theorem selection.

\begin{table*}[t]
\centering
\setlength{\tabcolsep}{3.5pt}
\begin{tabular}{lll@{\hspace{7pt}}ccc@{\hspace{9pt}}ccc}
\toprule
& & 
& \multicolumn{3}{c}{\textbf{Gemini-2.5-Flash}}
& \multicolumn{3}{c}{\textbf{Claude Sonnet 4.6}} \\
\cmidrule(lr){4-6} \cmidrule(lr){7-9}

\textbf{Setting} &
\textbf{Examples} &
\textbf{Dictionary} &
\shortstack{\textbf{1}\\\textbf{run}} &
\shortstack{\textbf{1 run}\\\textbf{+ retries}} &
\shortstack{\textbf{3 runs}\\\textbf{+ retries}} &
\shortstack{\textbf{1}\\\textbf{run}} &
\shortstack{\textbf{1 run}\\\textbf{+ retries}} &
\shortstack{\textbf{3 runs}\\\textbf{+ retries}} \\
\midrule

\textit{Ours} 
& Analogies 
& Analogy-derived 
& \textbf{48\%} & \textbf{74\%} & \textbf{86\%}
& \textbf{60\%} & \textbf{78\%} & \textbf{86\%} \\

(i) 
& Analogies 
& Full 
& 30\% & 68\% & 78\%
& 42\% & 62\% & 78\% \\

(ii) 
& Random 
& Random-derived
& 24\% & 54\% & 72\%
& 28\% & 60\% & 72\% \\

(iii) 
& Analogies 
& Random (size-matched)
& 32\% & 60\% & 78\%
& 36\% & 66\% & 74\% \\

\textit{Baseline} 
& Random 
& Full
& 22\% & 58\% & 72\%
& 28\% & 58\% & 78\% \\

\bottomrule
\end{tabular}

\caption{Controlled ablation of analogy retrieval and theorem-dictionary construction on 50 problems using Gemini-2.5-Flash and Claude Sonnet 4.6. Our full method performs best across all inference settings.}
\label{tab:ablation_dict}
\end{table*}

\noindent\textbf{Verifier feedback.} 
We evaluate the impact of verifier-guided retries. 
As shown in Table~\ref{tab:main_results_tab}, allowing retries consistently improves performance. 
For our analogy-based method, retries improve accuracy from 48\% to 68\% for o1, from 48\% to 74\% for Gemini, from 60\% to 78\% for Claude, and from 64\% to 96\% for GPT-5 ( \textit{ours}${1,1}$ and \textit{ours}${1,5}$).
The effect is even more pronounced for the base model, where verifier feedback increases accuracy from 10\% to 38\% for o1, from 22\% to 58\% for Gemini, from 28\% to 58\% for Claude, and from 44\% to 80\% for GPT-5 (\textit{base}${1,1}$ and \textit{base}${1,5}$).

\noindent\textbf{Multiple runs.}  
While RQ1 compared our full pipeline to the baseline, we now isolate the effect of multiple runs per method. Specifically, we compare \textit{ours}$_{3,5}$ vs. \textit{ours}$_{1,5}$ for our approach, and \textit{base}$_{3,5}$ vs. \textit{base}$_{1,5}$ for the base model. 
We find that additional runs consistently improve performance for both variants, yielding gains of  8\%-12\% for our method and 8\%-20\% for the base model (except GPT-5 under our method, which saturates at 96\%).

To conclude, \textbf{our method outperforms the baseline across all settings, for every model, with each component contributing to performance.}  


\subsection{Analysis.}
\label{subsec:further_analysis}
We provide additional analysis of our results, demonstrated on the o1 model; similar trends are observed for the other models.

\begin{figure}[t]
\includegraphics[width=0.39\textwidth]{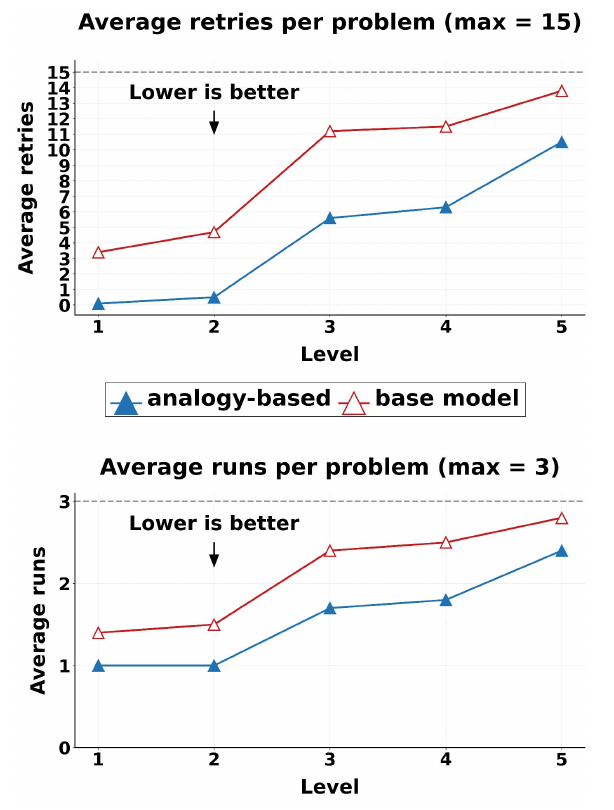}
\centering
\caption{
Average number of retries (top) and runs (bottom) per problem by difficulty level. The dashed line represents maximum allowed. Our analogy-based method consistently outperforms the base model (o1), with fewer retries and runs across all levels.
}
\label{fig:avg_runs_retries}
\end{figure}

\noindent\textbf{Effect of $m$ and $n$ parameters.}
Figure~\ref{fig:avg_runs_retries} shows the average number of retries (top) and runs (bottom) per problem by difficulty level. Dashed line represents maximum allowed, but the method stops early if it reaches a valid proof. Our method consistently requires fewer retries and runs than the base model across all levels.
On average, it uses 4.6 retries and 1.58 runs per problem, compared to 8.92 retries and 2.12 runs for the base model.

\noindent\textbf{Proof accuracy given correct numerical answers.}
To isolate \emph{proof-generation} failures from answer-generation failures, we evaluate whether models can produce a correct proof \emph{given that they found the correct numerical answer}.

We report numerical accuracy as pass@$k$, where $k$ indexes sequential, verifier-guided attempts, up to the complete inference budget $B{=}15$ (3 runs $\times$ 5 verifier-guided retries per run). For o1 with our analogy-based method, pass@1~=~90\% (45/50), pass@5~=~98\%, and pass@$B$~=~100\%. Our method achieves 100\% numerical pass@15, i.e., the correct numerical answer appears in at least one attempt for all 50 problems, and generates correct proofs for 40 of them (80\%). In contrast, the base model achieves 90\% numerical pass@15 (45/50 problems), but only 26 correct proofs (57.7\%, vs.\ 52\% overall).

Interestingly, the base model often finds the correct numeric answer but fails to produce a valid proof. We conjecture that the model has a partial understanding of the proof, yet struggles to express it precisely. By retrieving proofs from analogous problems, our method helps bridge this gap, improving both proof validity and overall answer accuracy (100\% vs.\ 90\% numerical pass@15).

\noindent\textbf{Stability of results.}
Our main evaluation uses 50 problems (10 per difficulty level, 1-5). To assess stability, we sample 10 additional problems per level and rerun our method, observing only 3\% average per-level variation. The baseline is similarly stable: Gemini-2.5-Flash changes from 72\% to 74\% (6\% per-level variation) and Claude Sonnet 4.6 from 78\% to 80\% (4\%). These results confirm the robustness of our evaluation sample.

\begin{figure}[t]
\includegraphics[width=0.39\textwidth]{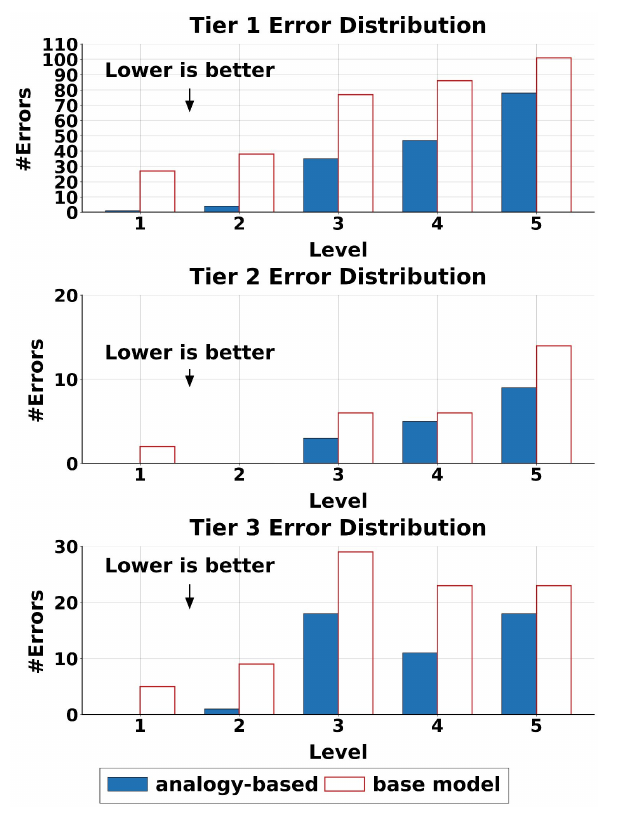}
\centering
\caption{Error distribution by tier for our method vs. (o1) base model. Our method reduces errors across all tiers, with tier-1 errors being most frequent in both. }
\label{fig:error_dist}
\end{figure}

\noindent\textbf{Error analysis.}
As detailed in Section~\ref{subsec:verifier}, we categorize verifier errors into three tiers. Figure~\ref{fig:error_dist} shows their distribution across difficulty levels.
Our method yields fewer errors than the base model across all tiers and levels.
Tier-1 syntax errors are the most prevalent, followed by Tier-3 errors, where the proof fails to establish the goal. Tier-2 errors, involving theorem calls with violated premises, are the least common. This trend also holds for the base model.
Tier-1 errors are frequent but easily corrected with verifier feedback, whereas Tier-2 errors reflect deeper geometric reasoning failures that often persist across retries. Thus, verification removes superficial errors, leaving reasoning failures as the main performance bottleneck.

\section{Related Work}

\noindent\textbf{Neural Models for Proof Generation.}
Prior work has used LLMs to generate logical and mathematical proofs \cite{tafjord2020proofwriter, kazemi2022lambada, yue2023mammoth}. More recently, LLM-based systems have achieved impressive results in formal theorem proving, including medal-level performance on International Mathematical Olympiad problems \cite{ren2025deepseek, hubert2026olympiad, chervonyi2025gold, tsoukalas2026advancing, achim2025aristotle}. These systems, including DeepSeek-Prover-V2, AlphaProof, AlphaGeometry 2, AlphaProof Nexus, and Aristotle, typically rely on specialized training or large-scale proof search. In contrast, we study how structured guidance and symbolic feedback can improve the proof-generation capabilities of general-purpose, off-the-shelf LLMs, without additional training.

Most closely related is Rango \cite{thompson2024rango}, which adaptively retrieves proofs and lemmas from the evolving Coq proof state for next-tactic prediction. In contrast, we retrieve structurally analogous problems from problem statements alone, once before generation, and use their complete proofs to guide whole-proof generation and select a compact set of relevant theorems, reducing context and search space.

\noindent\textbf{Neuro-symbolic Methods.} Neuro-symbolic approaches combine LLMs with symbolic tools. This strategy has been shown to be effective for some structured tasks \cite{szeider2024mcp,regin2024combining,mittal2024puzzlebench,wu2024can}.
Auto-formalization methods \cite{song2024towards, alhessi2025lemmanaid} such as LINC \cite{olausson2023linc} translate natural language into first-order logic using an LLM, followed by a symbolic theorem prover. These works focus on logical reasoning and are not well suited for mathematical or algebraic aspects of proofs.
One work in the domain of geometry is AlphaGeometry~\cite{trinh2024solving}, which combines symbolic deduction with a language model trained on 100M synthetic geometry examples to generate machine-verifiable proofs. In contrast, our method does not require training.


\noindent\textbf{Analogical Reasoning.}   
LLMs have been studied for analogical reasoning in diverse domains, including word analogies, narratives and processes, and visual scenarios.
see \citet{sultan2024parallelparc} for a survey. \citet{jobanputra2024teamsaarlst} show that retrieving structural analogies improves LLM-based data-to-text generation. In mathematics, \citet{yasunaga2023large} propose \emph{analogical prompting}, where the model generates its own analogous problems and solutions. In contrast, we retrieve structurally similar problems with verified proofs, avoiding potentially incorrect model-generated solutions.

\citet{zhou2025self} fine-tune models to transfer solutions between structurally similar problems, whereas our method requires no training and operates entirely at inference time. \citet{lee2025automatic} generate unlabeled numerical variants of math word problems as in-context examples, whereas we focus on formal proof generation, where such numerical variations offer little benefit.

\label{sec:related_work}

\section{Conclusions and Future Work}

We introduced a neuro-symbolic approach for proof generation, coupling LLMs with retrieval of analogous proofs and symbolic verification. The analogies help guide the LLM, and the verifier provides feedback on the generated proofs.

We evaluate our approach in Euclidean geometry. Experiments on the FormalGeo-7k dataset show substantial improvements in proof correctness over the base model across four leading LLMs from three families, with each component contributing to the gains. It also reduces cost by narrowing the theorem dictionary via analogous proofs.
These results demonstrate the potential of neuro-symbolic methods for mathematical reasoning.


In the future, we plan to extend our approach beyond geometry to other 
areas, where automated proof verification could help validate complex derivations and ensure the consistency of models. We also see an interesting use case in education, where analogical retrieval can surface similar solved problems to guide students, and a symbolic verifier can offer targeted hints and feedback. 

We also aim to explore more expressive forms of formal verification, such as model checking with temporal logics (LTL and CTL), enabling reasoning about dynamic systems that evolve over time, as well as more sophisticated mechanisms for analogy retrieval and 
structural similarity.

We hope this work inspires future research on neuro-symbolic systems that combine the flexibility of LLMs with the precision of formal reasoning, as a step toward more reliable AI systems for domains where correctness is crucial.



\section*{Limitations}

\subsection*{General Limitations.}
\begin{itemize}
\item \textbf{Generalization beyond geometry.}
Our experiments focus on Euclidean geometry, and results may differ in other formal domains, which may involve different proof structures, theorem types, or symbolic representations.
While our pipeline is designed to be adaptable to any domain that supports SMT-based verification, further evaluation is needed to confirm its effectiveness beyond geometry.
\end{itemize}

\subsection*{Analogy Retrieval Limitations.}

\begin{itemize}

    \item \textbf{Abstraction schema for analogy retrieval.} 
    We use a simple abstraction method that replaces entity names and numbers with placeholders. 
    While effective in many cases, this schema may miss deeper structural nuances or semantic relationships, leading to suboptimal or misleading analogies. 
\end{itemize}

\subsection*{Verifier Limitations.}

\begin{itemize}
    \item \textbf{No support for inverse trigonometric.}
    A key limitation of our verification system is that the Z3 theorem prover does not support inverse trigonometric functions. While we implemented workarounds for direct functions such as sine and cosine, the system cannot verify cases requiring inverse operations (e.g., computing $\theta$ from $\cos(\theta) = 0.5$). This limits our ability to validate proofs where angle measures must be derived from trigonometric values. Future work could address this by integrating custom decision procedures for inverse trigonometric reasoning.
\end{itemize}

\subsection*{LLM Limitations.}

\begin{itemize}
  \item \textbf{No access to diagrams.}
  Our LLM uses only formal and textual descriptions, without visual inputs.
  Although our dataset includes rich symbolic annotations, covering geometric entities, relationships, constructions, and extended conditions, some errors arise due to the system's lack of access to visual diagrams. For instance, when reasoning about arcs, the model may incorrectly infer that $\mathrm{BOD} + \mathrm{BOA} = \mathrm{AOD}$ instead of the correct relation $\mathrm{BOD} = \mathrm{BOA} + \mathrm{AOD}$. 
  Such mistakes are less frequent in  reasoning about line, where properties like collinearity are explicitly annotated. In some cases, ambiguity in notation further complicates interpretation; for example, the angle name $\angle DOB$ could refer to the angle itself or its complement, depending on context.
  
  On the other hand, we note that a recent work \cite{zhang2024mathverse} has shown that multimodal LLMs often struggle with the visual aspects of math problems, often performing better without the images.




    \item \textbf{LLMs are sensitive to prompt phrasing.}
    LLMs are sometimes sensitive to small changes to the prompts.
\end{itemize}

\label{sec:limitations}

\section*{Ethical Considerations}

\noindent\textbf{Dataset.}  
To protect privacy and ensure proper anonymization, we removed the names of individual annotators from the subset of problems used in our experiments and shared on GitHub.

\noindent\textbf{Use of AI Assistants.}  
We used Claude Sonnet 4.6 for coding assistance and GPT-5 for language editing and rephrasing. All AI-generated outputs were reviewed and edited by the authors to ensure alignment with our design goals and original intent.

\paragraph{Acknowledgements}

We thank the members of HyadataLab and NLP-HUJI for their valuable feedback, as well as the anonymous reviewers for their constructive comments. This work was supported by the European Research Council (ERC) under the European Union’s Horizon 2020 research and innovation programme (grant no. 852686, SIAM).

\bibliography{custom}

\appendix
\label{appendix}

\section{Analogous Problems Retrieval}

We trained a neural network to predict proof similarity between two problems, using the Jaccard similarity of their proofs as the target. The model inputs a 3D feature vector capturing structural and semantic similarities: (1) Jaccard over abstract constructions, (2) Jaccard over abstract conditions, and (3) goal similarity -- computed from abstracted problem representations.
The model has three fully connected layers: 3→128 (ReLU), 128→64 (ReLU), and 64→1 (linear). It was trained for 10 epochs with batch size 32 using MSE loss and the Adam optimizer (learning rate 0.001).

\label{appendix:analogous_problems_retrieval}

\section{LLM Proof Generation}

See Figure~\ref{tab:llm_system_prompt} for the system prompt provided to the LLMs, and Figure~\ref{tab:llm_few_shot_prompt} for few-shot examples of analogous problems and their solutions (including proof, and answer).

\begin{figure*}[t]
\centering
\fbox{%
\begin{minipage}{0.99\textwidth}
\small  

You are a mathematician expert in solving geometry problems. \\
Your task is to solve Problem B by constructing the correct sequence of theorems (THEOREM\_SEQUENCE) to form its proof. \\  

\textbf{Inputs:} \\
You are given: \\
Problem B, which you need to solve. \\
Five analogous problems (A1, A2, A3, A4, A5) that have similar proof structures. These are provided to help guide your approach to solving Problem B. \\
Geometry Theorem Dictionary (GDL): A dictionary containing various geometry theorems, each with its premise and conclusion. All theorems in the THEOREM\_SEQUENCE must be selected from this dictionary. \\
GDL\_DICTIONARY: \\
\{GDL\} \\
For each problem, you are provided the following data: \\
DESCRIPTION: A textual description of the problem. \\
CONSTRUCTION\_CDL: The problem's construction in Condition Declaration Language (CDL). \\
TEXT\_CDL: The text of the problem in CDL. \\
GOAL\_CDL: The goal of the problem in CDL. \\
CONSTRUCTION\_CDL\_EXTENDED: An extended version of CONSTRUCTION\_CDL. \\
SYMBOLS\_AND\_VALUES: Symbols with corresponding values, in the format (predicate;symbol;value). \\
EQUATIONS: Equations related to solving the problem, in the format (equation). \\
GOAL\_SYMBOL: The symbol you are trying to solve for, in the format (symbol). \\
ANSWER: The calculated final answer, in the format (answer). \\
THEOREM\_SEQUENCE: The sequence of theorems used in the proof, formatted as: \\
step\_id <step\_id>; <theorem>; <premise>; <conclusion> \\
step\_id <step\_id>; <theorem>; <premise>; <conclusion> \\ 

\textbf{Your Task:} \\
You need to solve Problem B by constructing the correct THEOREM\_SEQUENCE, which should consist of theorems from the GDL. Ensure that each selected theorem logically follows the previous one and contributes to the goal of solving Problem B. \\ 

\textbf{Output Format:} \\
Your response must contain the following: \\
EQUATIONS: <equation> <equation> ... \\
GOAL\_SYMBOL: <symbol> \\
ANSWER: <answer> \\
THEOREM\_SEQUENCE: \\
<step\_id>; <theorem>; <premise>; <conclusion> \\
<step\_id>; <theorem>; <premise>; <conclusion> \\ 

\textbf{Important Notes for the THEOREM\_SEQUENCE:} \\
Do not include the words "theorem", "premise", or "conclusion". Your sequence should only contain the step ID, theorem name, premise, and conclusion. \\
Use the exact theorem names and formats provided in the GDL. \\
Start with step\_id = 1 and increment sequentially. \\
When referring to angles, use three letters (e.g., ABC for the angle at B). Be mindful of the order, as ABC is different from ACB. For polygons, list all distinct points in clockwise or counterclockwise order. For example, a polygon with points FGE can also be referred to as GEF or EFG. \\

\textbf{Example of Correct THEOREM\_SEQUENCE Format:} \\
1; angle\_addition(1,BFE,EFG); \\
Angle(BFE)\&Angle(EFG)\&Angle(BFG); \\
\texttt{["Equal(MeasureOfAngle(BFG),Add(MeasureOfAngle(BFE),MeasureOfAngle(EFG)))"]} \\
2; triangle\_property\_angle\_sum(1,DFC); Polygon(DFC); \\
\texttt{["Equal(Add(MeasureOfAngle(DFC),MeasureOfAngle(FCD),MeasureOfAngle(CDF)),180)"]} \\

\textbf{Reminder:} \\
Ensure that the theorems you select come from the GDL, follow the correct format, and use the proper arguments (e.g., angle order and polygon points). 
Also, pay attention to the specific variation of the theorem (e.g., 1 for the first variation, 2 for the second, etc.).

\end{minipage}
}
\caption{System prompt given to the o1 LLM for generating a geometry proof for a target problem B. The model is provided with all details about the task, as well as a Geometry Theorem Dictionary (GDL) and five analogous problems (A1–A5), which are supplied as few-shot examples (see Figure~\ref{tab:llm_few_shot_prompt}).}

\label{tab:llm_system_prompt}
\end{figure*}

\begin{figure*}[t]
\centering
\fbox{%
\begin{minipage}{0.99\textwidth}
\small

\textbf{Inputs for Problem A1:} \\
\textbf{DESCRIPTION:} \\
As shown in the diagram, Div(LengthOfLine(AD)=LengthOfLine(DF)), Div(LengthOfLine(DF)=LengthOfLine(FB)), AG=15, \\
DE is parallel to BC, DE is parallel to FG, FG || BC. Find the length of line CE. \\

\textbf{CONSTRUCTION\_CDL:} \\
Shape(AD,DE,EA), ... \\
Collinear(ADFB), ... \\

\textbf{TEXT\_CDL:} \\
Equal(Div(LengthOfLine(AD),LengthOfLine(DF)),3/2), ... \\
ParallelBetweenLine(DE,BC), ... \\

\textbf{GOAL\_CDL:} \\
Value(LengthOfLine(CE)) \\

\textbf{CONSTRUCTION\_CDL\_EXTENDED:} \\
Shape(DE,EA,AD), ... \\
Collinear(BFDA), ...\\
Point(A), ... \\
Line(AB), ... \\
Angle(ADF), ... \\
Polygon(ADE), ... \\
ParallelBetweenLine(CB,ED), ... \\

\textbf{SYMBOLS\_AND\_VALUES:} \\
LengthOfLine(AG);ll\_ag;15, ... \\

\textbf{Outputs for Problem A1:} \\
\textbf{EQUATIONS:} \\
ll\_ad/ll\_df - 3/2, ... \\

\textbf{GOAL\_SYMBOL:} \\
ll\_ce \\

\textbf{ANSWER:} \\
9 \\

\textbf{THEOREM\_SEQUENCE:} \\
... \\

\vspace{1em}
\textbf{Inputs for Problem A2:} \\
...

\vspace{1em}
\textbf{Inputs for Problem A3:} \\
...

\vspace{1em}
\textbf{Inputs for Problem A4:} \\
...

\vspace{1em}
\textbf{Inputs for Problem A5:} \\
...

\vspace{1em}
\textbf{Outputs for Problem B:}
\end{minipage}
}
\caption{Few-shot prompt shown to the o1 LLM, consisting of five analogous problems (A1–A5) along with their inputs and solutions -- including the final answer and theorem sequence used in the proof. \textbf{Note:} ``...'' indicates omitted content for brevity.}

\label{tab:llm_few_shot_prompt}
\end{figure*}

\label{appendix:LLM_proof_generation}

\section{Verifier Implementation Details}

\paragraph{Geometric reasoning module.}
We represent each problem state using a structured system of Python classes that encodes geometric facts, such as point orderings, segment lengths, and angle relationships. To enhance robustness and flexibility, we apply normalization procedures that map geometric objects to canonical, direction-agnostic forms. For example, treating  $\angle$ ABC and  $\angle$ CBA as equivalent.

\paragraph{Algebraic constraint solver.}
We use the Z3 Theorem Prover, a state-of-the-art SMT solver, to represent and enforce algebraic constraints implied by geometric statements. When a proof step establishes a fact like collinearity of points A, B, C, and D, the system automatically generates angle equality constraints (e.g., $\angle XAB = \angle XAC = \angle XAD$ for any reference point X).

\paragraph{Handling of Trigonometric Expressions.}
Z3 does not natively support trigonometric functions. To work around this limitation, we:
\begin{itemize}
  \item Introduce symbolic variables for trigonometric expressions (e.g., \texttt{cos\_ABC}).
  \item Add angles referenced by trigonometric terms to the system’s internal constraint graph.
  \item Evaluate goals involving trigonometric expressions either by matching against known values or by reasoning symbolically if exact evaluation is not possible.
\end{itemize}

\paragraph{Verifier Examples}

\begin{figure*}[!t]
    \centering
    \fbox{%
    \begin{minipage}{0.99\textwidth}
    \ttfamily

    \textbf{Tier 1 (Syntax Violation)} \\
    \hrule\vspace{0.5em}
    Theorem: parallel\_property\_ipsilateral\_internal\_angle(1,GA,HD) \\
    You output the following premises: ParallelBetweenLine(GA,HD)\&Line(AD) \\
    But the correct premises: ParallelBetweenLine(GA,HD)\&Line(GH) \\

    \textbf{Tier 2 (Premise Violation)} \\
    \hrule\vspace{0.5em}
    - Error: You tried to use theorem: right\_triangle\_judgment\_angle(1,BCD);\\
    Polygon(BCD)\&Equal(MeasureOfAngle(BCD),90);['RightTriangle(BCD)'] \\ 
    
    Missing premise: Angle measure 90° for triangle BCD is not established in the premise.\\
    Details: Premise provided: Polygon(BCD)\&Equal(MeasureOfAngle(BCD),90) \\
    - Available premises: \\
      Perpendicular Lines: AD, BD \\
      Collinear Points: ADC \\
      Polygons: ABC, ABD, ACB, ADB, BCD, BDC \\ 
    - Theorems related to the goal: \\ Step 1 - right\_triangle\_judgment\_angle(1, BCD): RightTriangle(BCD) \\ 
    - Solver constraints directly related to this goal: \\ |AB| = y \\ 
    Please fix the proof. \\[2ex]

    \textbf{Tier 3 (Goal Not Reached)} \\
    \hrule\vspace{0.5em}
    - Goal: measure of angle ADB \\
    - Model answer: 55.0 \\
    - Error: Your proof doesn't uniquely determine the value. You need to use additional theorems to constrain the value. \\
    - Available premises: \\
      Collinear Points: AOB, BCD \\
      Cocircular Points: A on circle O, AB on circle O, ABC on circle O, AC on circle O, \\
      B on circle O, BC on circle O, C on circle O \\
      Circles: O center: O \\
      Circle Diameters: AB diameter of O \\
      Tangent Lines: DA tangent to O \\
      Polygons: ABD, ADB, ADCO, BCO, BOC, COAD, DCOA, OADC \\
    - Theorems related to the goal: None found that constrain this goal \\
    - Solver constraints directly related to this goal: 
$\angle ADB \leq 180$, \quad 
$\angle ADB > 0$, \quad 
$\angle ADC = \angle ADB$ \\

    \end{minipage}
    }
    \caption{
        Examples of verifier errors across three tiers. \textbf{Tier 1}: syntax violations in theorem calls. \textbf{Tier 2}: missing or undefined premises. \textbf{Tier 3}: under-constrained reasoning that fails to derive the goal.
    }
    \label{tab:tiered_verifier_errors}
\end{figure*}

Figure~\ref{tab:tiered_verifier_errors} presents a representative example of error outputted by the verifier, from each error tier.

\label{appendix:verifier_details}

\section{RQ1: Performance Gains }

\begin{figure}[t]
\includegraphics[width=0.48\textwidth]{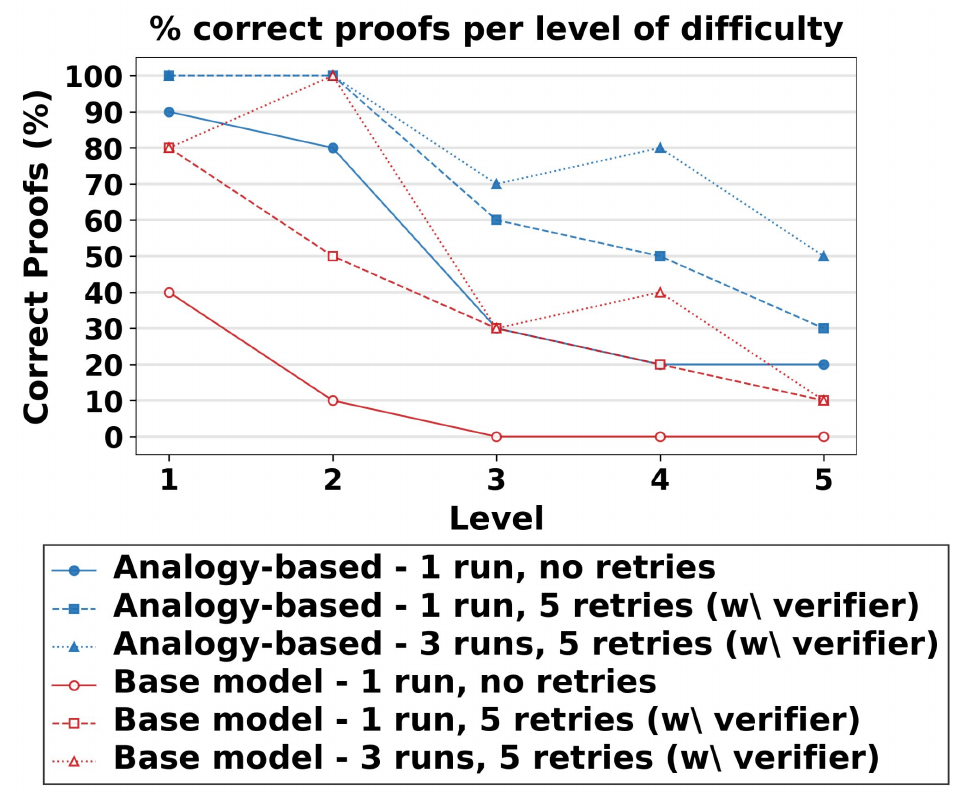}
\centering
\caption{\% correct proofs per level of difficulty (50 samples, 10 per level). 
Our analogy-based method outperforms the o1 base model (non-analogy) in all settings. Analogy retrieval, verifier feedback, and multiple runs each significantly contributes to performance.
Our full pipeline (blue triangle) outperforms the baseline in every level, reaching an average aggregated accuracy of 80\%. Even without multiple runs (blue square), performance remains strong at 68\%, far exceeding the 10\% of the base model baseline (red hollow circle).
}
\label{fig:experiments_result}
\end{figure}

\begin{figure}[t]
\includegraphics[width=0.48\textwidth]{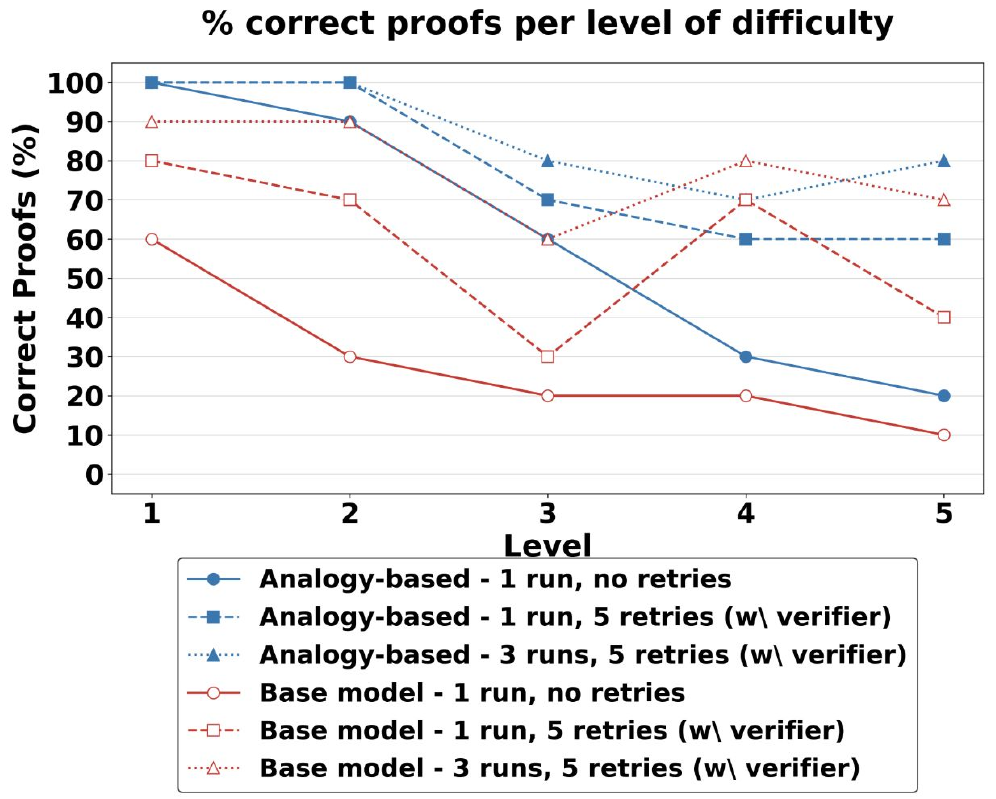}
\centering
\caption{\% correct proofs per level of difficulty (50 samples, 10 per level). 
Our analogy-based method outperforms the Claude Sonnet-4.6 base model (non-analogy) in all settings. Analogy retrieval, verifier feedback, and multiple runs each significantly contributes to performance.
Our full pipeline (blue triangle) outperforms the baseline in every level, reaching an average aggregated accuracy of 86\%. Even without multiple runs (blue square), performance remains strong at 78\%, far exceeding the 28\% of the base model baseline (red hollow circle).
}
\label{fig:experiments_result_claude}
\end{figure}

\begin{figure}[t]
\includegraphics[width=0.48\textwidth]{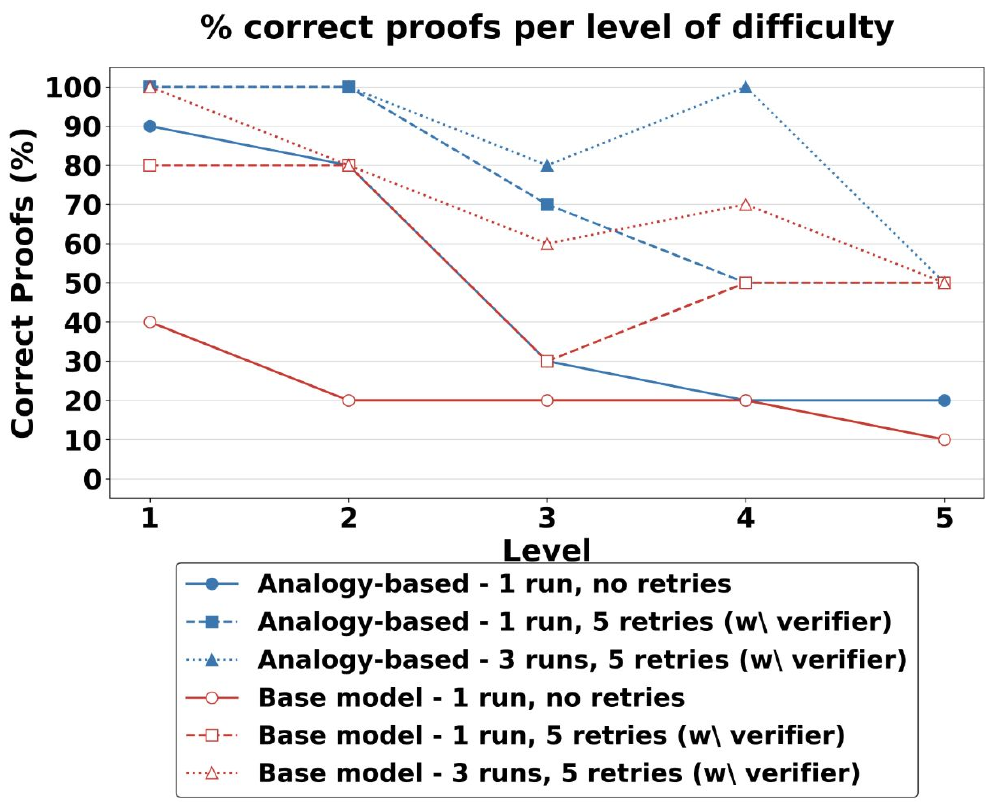}
\centering
\caption{\% correct proofs per level of difficulty (50 samples, 10 per level). 
Our analogy-based method outperforms the Gemini-Flash-2.5 base model (non-analogy) in all settings. Analogy retrieval, verifier feedback, and multiple runs each significantly contributes to performance.
Our full pipeline (blue triangle) outperforms the baseline in every level, reaching an average aggregated accuracy of 86\%. Even without multiple runs (blue square), performance remains strong at 74\%, far exceeding the 22\% of the base model baseline (red hollow circle).
}
\label{fig:experiments_result_gemini}
\end{figure}

\begin{figure}[t]
\includegraphics[width=0.48\textwidth]{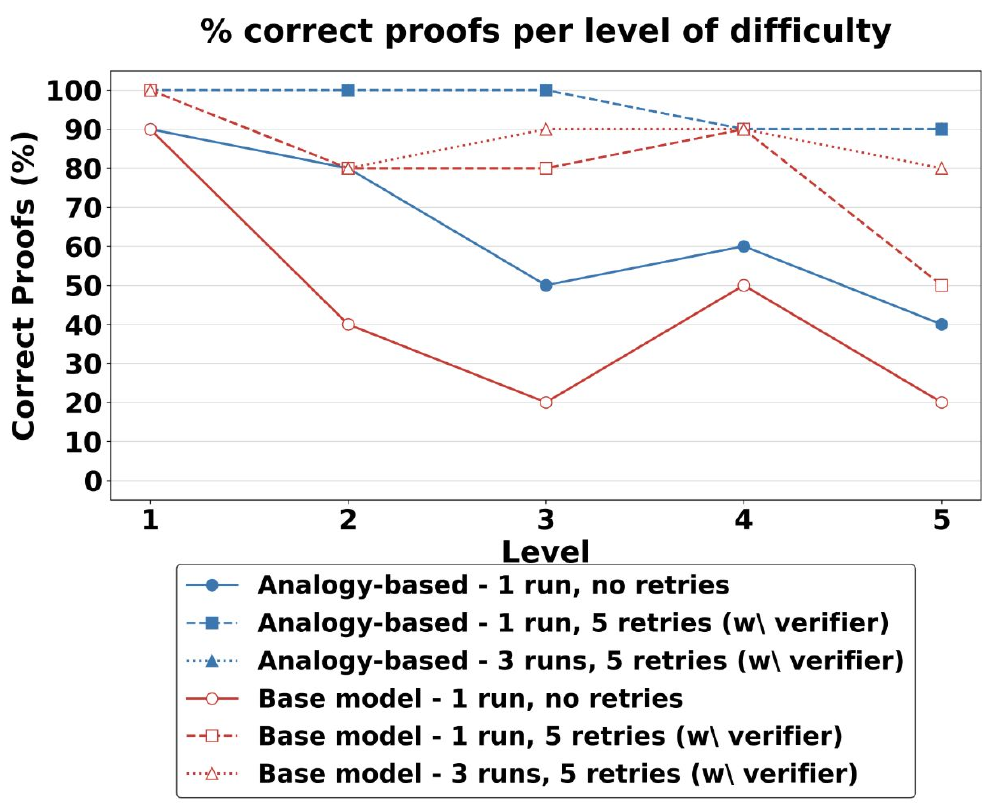}
\centering
\caption{\% correct proofs per level of difficulty (50 samples, 10 per level). 
Our analogy-based method outperforms the GPT-5 base model (non-analogy) in all settings. Analogy retrieval, verifier feedback, and multiple runs each significantly contributes to performance.
Our full pipeline (blue triangle) outperforms the baseline in every level, reaching an average aggregated accuracy of 96\%. Even without multiple runs (blue square), performance remains stable at 96\%, far exceeding the 44\% of the base model baseline (red hollow circle).
}
\label{fig:experiments_result_gpt5}
\end{figure}

We provide additional details on the results and statistical analyses, focusing on the o1 model; similar trends are observed across other models.

Figure~\ref{fig:experiments_result} shows the results on o1 model for the different settings. 
We start by comparing our full pipeline (blue triangle: analogy + verifier, multiple runs with retries), to the most basic baseline (red hollow circle: non-analogy, no verifier, first run, no retries). Our full pipeline consistently outperforms the baseline at every level, achieving an average aggregated accuracy of 80\%. In contrast, the base model achieves an average of only 10\% (with 0\% accuracy on levels 3 and above).
This difference is statistically significant with $p=$5.82e-11 in the McNemar test at the 0.05 level (our method solved all problems the base model did, plus 35 others).

To isolate the contribution of multiple runs, we also evaluated our model after a single run (blue square). This setting also substantially outperformed the base-model baseline, achieving an overall accuracy of 68\%,
with $p=$3.73e-09 in the McNemar test at the 0.05 level. 
Thus, we conclude that our method significantly boosts the base model’s success rate in generating correct proofs, even without multiple runs.
See Figures~\ref{fig:experiments_result_gemini}, \ref{fig:experiments_result_claude}, and \ref{fig:experiments_result_gpt5} for the corresponding results on Gemini-Flash-2.5, Claude Sonnet-4.6, and GPT-5, respectively. 
\label{appendix:results_rq1}

\section{RQ2: Ablations}

We provide additional details on the results and statistical analyses, focusing on the o1 model; similar trends are observed across other models.

We now assess the contribution of the different components of our pipeline: The analogy retrieval, the verifier, and the multiple runs.

\noindent\textbf{Analogy retrieval.}
To measure the effect of analogy retrieval, we compare our method to the base model (non-analogy), under the same three settings:
(1) first run, no retries,
(2) first run, with retries (verifier),
(3) multiple runs, with retries (verifier). 

In setting (1) our method achieved an overall accuracy of 48\%, substantially outperforming the baseline's 10\% (blue circle vs. red hollow circle), $p=$3.81e-06 in the McNemar test at the 0.05 level. 
In setting (2), our method achieved 68\% overall accuracy, while the base model reaches 38\% (blue square vs. red hollow square), $p=$6.10e-05 in the McNemar test at the  0.05 level. 
In setting (3) we observe a 80\% overall accuracy for our method vs. 52\% for the base model (blue triangle vs. red hollow triangle). This improvement is also statistically significant, with $p=$1.22e-04 in the McNemar test at the  0.05 level. 
That is, analogy retrieval boosts model performance across all settings.

\noindent\textbf{Disentangling analogy retrieval and theorem dictionary construction.}
The comparison above jointly varies two factors: whether examples are retrieved analogies or random problems, and whether the theorem dictionary is analogy-derived or full. To isolate each factor's individual contribution, we evaluate three additional configurations on Gemini-Flash-2.5 and Claude Sonnet-4.6, using the same 50 evaluation problems and inference protocol (Table~\ref{tab:ablation_dict}): (i) retrieved analogies with the full theorem dictionary, isolating the effect of analogy retrieval alone; (ii) random examples with a theorem dictionary narrowed from those random examples, isolating the effect of dictionary narrowing without analogy retrieval; and (iii) retrieved analogies with a random, size-matched theorem dictionary, isolating the effect of the analogy-derived theorem selection specifically.

Our complete method consistently achieves the best performance across all evaluation settings, and is the top-performing configuration for both models under a single attempt, verifier-guided retries, and the full inference budget.

The intermediate ablations further validate the contribution of analogy retrieval. Both setting (i) (retrieved analogies + full dictionary) and setting (iii) (retrieved analogies + random size-matched dictionary) consistently outperform setting (ii) (random examples + narrowed dictionary), showing that retrieving analogous proofs provides useful guidance even before considering theorem-dictionary construction. Compared with the baseline, settings (i) and (iii) improve performance in nearly all evaluation settings, with the largest gains on the first attempt (e.g., +8 and +10 points for Gemini, +14 and +8 points for Claude).

Finally, comparing our method with settings (i) and (iii) isolates the contribution of the analogy-derived theorem dictionary. Replacing the analogy-derived dictionary with either the full theorem dictionary (setting (i)) or a random size-matched dictionary (setting (iii)) consistently reduces accuracy, demonstrating that the improvement comes from selecting theorems that are specifically relevant to the retrieved analogies, rather than simply reducing prompt size. This is consistent with Table~\ref{tab:theorems_coverage}, where the analogy-derived theorem dictionary retains the complete target-proof theorem set for more problems than a random-example-derived dictionary of comparable size, while reducing the prompt from approximately 18K to 2.5K tokens on average.

\noindent\textbf{Verifier feedback.}
We now measure the impact of retries following feedback.
As shown in Figure~\ref{fig:experiments_result}, allowing retries consistently improves results across all difficulty levels. For our analogy-based method, 
retries (blue square) yields an average gain of additional 20\% over no retries (blue circle), with gains ranging from 10\% to 30\% at different levels.
The improvement is even more pronounced for the base model (red hollow square vs.\ red hollow circle), where verifier feedback results in an average gain of 28\%, ranging from 10\% to 40\% per level.

\noindent\textbf{Multiple runs.}  
In RQ1, we evaluated the effect of multiple runs when comparing our full method to the base model. We now analyze the impact of multiple runs for the same method (blue triangle vs.~square, red hollow triangle vs.~square).
We find that additional runs improve performance for both our method and the base model. Notably, for our method multiple runs help more on harder problems (gains of 20-30\% for levels 4-5), where the potential for improvement is greater.


\smallskip

To conclude, our method outperforms the baseline across all settings, with each component contributing to performance.  Although the verifier alone does enhance the base model’s results, analogy retrieval supplies stronger initial proof candidates for the verifier to correct.  


\label{appendix:results_rq2}

\section{Stability of Results}

\begin{figure*}[t]
\includegraphics[width=0.99\textwidth]{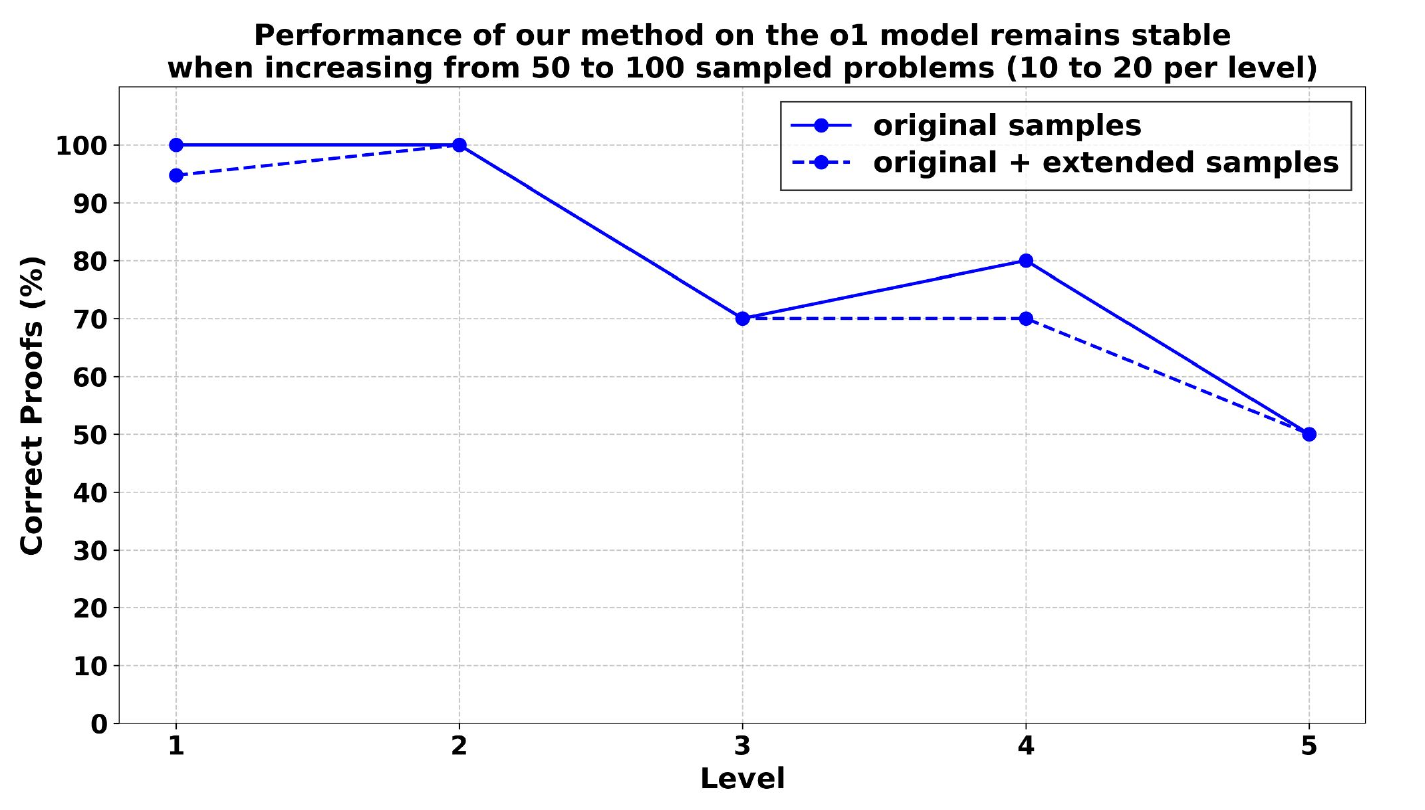}
\centering
\caption{
Accuracy remains stable when increasing the sample size from 50 to 100 (10 additional problems per level), with an average variation of only 3\% per level. Per-level differences range from 0\% (levels 2, 3, and 5) to 5\% (level 1), and 10\% (level 4). We conclude our results are stable.
}
\label{fig:stability_of_results}
\end{figure*}

Figure~\ref{fig:stability_of_results} shows the performance of our method on both the original 50 samples and the extended set of 100 samples, broken down by level.
As shown, performance remains unchanged in levels 2, 3, and 5, while levels 1 show 5\% difference, and level 4 4 show a 10\% difference. Overall, the average difference across levels is 3\%. We conclude our results are stable.

We additionally repeat this stability check for the baseline (non-analogy) variant, on Gemini-Flash-2.5 and Claude Sonnet-4.6, given its substantially higher inference cost relative to our method (the baseline prompts with the full theorem dictionary rather than our analogy-derived, narrowed one). Baseline accuracy is similarly stable when increasing the evaluation set from 50 to 100 problems: for Gemini-Flash-2.5, overall accuracy increases only from 72\% to 74\%, with an average per-level variation of 6\%; for Claude Sonnet-4.6, overall accuracy increases only from 78\% to 80\%, with an average per-level variation of 4\%. These results confirm that our conclusions are robust to the evaluation sample size for both the proposed method and the baseline.
\label{appendix:stability_of_results}


\end{document}